# Automated Driving Maneuvers under Interactive Environment based on Deep Reinforcement Learning

Pin Wang, Ching-Yao Chan, Hanhan Li
University of California, Berkeley
{pin_wang, cychan, h_li}@berkeley.edu

## ABSTRACT

Safe and efficient autonomous driving maneuvers in an interactive and complex environment can be considerably challenging due to the unpredictable actions of other surrounding agents that may be cooperative or adversarial in their interactions with the ego vehicle. One of the state-of-the-art approaches is to apply Reinforcement Learning (RL) to learn a time-sequential driving policy, to execute proper control strategy or tracking trajectory in dynamic situations. However, direct application of RL algorithms is not satisfactorily enough to deal with the cases in the autonomous driving domain, mainly due to the complex driving environment and continuous action space. In this paper, we adopt Q-learning as our basic learning framework and design a unique format of the Q-function approximator that consists of neural networks to handle the continuous action space challenge. The learning model is present in a closed form of continuous control variables and trained in a simulation platform that we have developed with embedded properties of real-time vehicle interactions. The proposed algorithm avoids invoking an additional actor network that learns to take actions, as in actor-critic algorithms. At the same time, some prior knowledge of vehicle dynamics is also fed into the model to assist learning. We test our algorithm with a challenging use case - lane change maneuver, to verify the practicability and feasibility of the proposed approach. Results from accumulated rewards and vehicle performance show that RL vehicle agents successfully learn a safe, comfort and efficient driving policy as defined in the reward function.

## INTRODUCTION

Reinforcement Learning (RL) has been applied in robotics for decades (*1*) and has gained popularity due to the development in deep learning. In some recent studies, it has been applied for learning 3D locomotion tasks such as bipedal locomotion and quadrupedal locomotion (*2*), and robot arm manipulation tasks such as stacking blocks (*3*). Google DeepMind also showed the power of RL for learning to play Atari 2600 games (*4*) and Go games (*5*). In these applications, either the operation environment is simple, e.g. only the ego agent maneuves in the robotic cases, or the action space is taken as discrete, e.g. in some of the computer games.

In regard to the driving tasks for automated vehicles, the situation is totally different because vehicles are required to operate safely and efficiently in an extremely dynamic and complicated driving environment. Tough challenges frequently arise in automated driving domains. For example, the vehicle agent needs not only to avoid collisions with objects either in motion or stationary, but also to coordinate with surrounding vehicles so as not to disturb the traffic flow significantly when it executes a maneuver, e.g. a lane change action. The most challenging part is the reaction from surrounding vehicles which may be highly unpredictable. For example, in a lane change maneuver the trailing vehicle in the target lane may respond cooperatively (e.g. decelerate or change lane to yield to the ego vehicle) or adversarially (e.g. accelerate to deter the ego vehicle from cutting into its course). Thereby, training a vehicle for safe and effective driving under an interactively dynamic environment is of great importance for the deployment of automated driving systems.

Vehicle driving maneuvers are essentially time sequential problems where the completion of a task (e.g. a lane change/ramp merge) involves a sequence of actions taken under a series of states, and that the action at the current time has a cumulative impact on the ultimate goal of the task (e.g. a safe and comfortable lane change/ramp merge). Reinforcement Learning can take into account the interaction between the learning agent and the environment in the problem formulation, and is appropriately applicable in finding suitable control strategies.

There have been some efforts in applying reinforcement learning to automated vehicles (*6*) (*7*) (*8*), however, in some of the applications the state space or action space are arbitrarily discretized to fit into the RL algorithms (e.g. Q-learning) without considering the specific characteristics of the studied cases. Simplified discretization always leads to the loss of full and appropriate representation in the continuous space. Some policy gradient based methods are alternative ways for solving continuous action space problems but need the design of an action network which often suffers a hard time in learning a good policy from pure neural network design. To avoid such dillema, in this work we propose a novel Q-function approximator on the basis of Q-learning to find optimal driving policies in continuous state space and action space, and put some prior knowledge of vehicle motion mechanism into the learning model for fast learning. In particular, we use lane change scenario as an illustrative use case to explain our methodology.

The rest of the paper is organized as follows. Related work is described in Section II. Section III introduces the methodology and details of our application case. Validation of the proposed approach based on a simulation platform follows in Section IV. Concluding remarks and discussion of future studies are given in the last section.

## RELATED WORK

In the automated vehicle field, a vast majority of self-driving algorithms are based on traditional methods that rely on predefined rules or models to explicitly construct a logic architecture of the mechanisms on how vehicles behave under different situations. For example, in (*9*), a virtual trajectory reference was established with a polynomial function for each moving vehicle, and a bicycle model was used to estimate vehicle positions by following the calculated trajectory. In (*10*), a number of way points with information acquired from Differential Global Positioning System and Real-time Kinematic devices were used to generate a guidance path for the vehicle to follow. Such approaches can work well in predefined situations or within the model limits, however, a limitation is the lack of flexibility under dynamic or emergency driving conditions.

Some optimization-based approaches, e.g. MPC, which have been successfully applied for trajectory planning under constrained conditions (*11*), also suffer from the aforementioned limitations in dealing with extremely complicated driving conditions. Besides, the optimization criteria for real-world driving problems may become too complex to be explicitly formulated for all scenarios, particularly when vehicles' behaviors are stochastic.

Reinforcement learning algorithms, with the capability of dealing with time-sequential problems, can seek optimal policies by learning from trials and errors. The concept was recently introduced in automated driving related studies. Yu et al. (*6*) explored the application of Deep Q-Learning methods, with some modifications like prioritized experience replay and double Deep Q-Network, on the control of a simulated car in JavaScript Racer. They discretized the action space into nine actions, and found that the vehicle agent can learn turning operations in large-scale regions with no cars on the raceway, but cannot perform well in obstacle avoidance. Ngai et al. (*7*) incorporated multiple goals, i.e. destination seeking and collision avoidance, into the reinforcement learning framework to address the autonomous overtaking problem. They converted continuous sensor values into discrete state-action pairs with the use of a quantization method and took into account some of the responses from other vehicles. In these applications, the action space was treated as discrete and only simple interaction with the surrounding environment was considered.

Sallab et al. (*8*) moved further to explore the application of reinforcement learning on both discrete action space and continuous action spaces, still under some simple environment interactions. They proposed two corresponding learning structures, deep deterministic actor-critic and deep Q-learning, for problems in these two domains. Lane keeping scenarios were used as application case and simulated in an Open Racing Car Simulator (TORCS). Comparison results showed that vehicles performed more smoothly under continuous action design than the other approach with discrete action design.

Q-learning is simple and efficient as it is model free and only needs a value function network, but it is generally applied to discrete action space problems. If we can design a Q-function approximator which encodes the continuous action space to corresponding Q-values, it will help avoid involving a complicated policy network design as in policy gradient methods (*12*) (*13*), and at the same time ensures the efficient learning ability of the Q-learning. With this regard, we design a quadratic format of Q-function approximator with the idea of normalized advantage functions (NAF) as mentioned in (*14*), and use a two-stage training process for stable and efficient learning of a safe, comfort and efficient driving maneuver.

## METHODOLOGY
In this section, we first introduce the quadratic Q-function approximator built on a normalized advantage function, and then describe the driving maneuver control structure. To avoid the coupling impacts from lateral and longitudinal directions on the learning ability, the work reported in this paper uses RL to learn one dimensional action, i.e. the lateral control. The longitudinal action is determined by an adapted Intelligent Driver Model (IDM) which is well-developed for car-following behaviors. The methodology will be revised to accommodate two-dimensional action space in upcoming studies.

### Quadratic Q-function Approximation
*Q-function Based on Normalized Advantage Function*

In a typical reinforcement learning problem, an agent takes an action $a \in A$ based on the current state $s \in S$, and receives an immediate reward $r \in R$ at each time step. The state space $S$ and action space $A$ can be either discrete or continuous depending on the problems. The goal of the reinforcement learning is to find an optimal policy $\pi^*: S \rightarrow A$, so that the total expected return $G$ of the discounted immediate rewards accumulated over the course of the task is maximized.

For a given policy $\pi$ with parameters $\theta$, a Q value function is used to estimate the total reward from taking action $a_t$ in a given state $s_t$ at time $t$, Equation ($1$). A value function is to estimate the total reward from state $s_t$, Equation ($2$). The advantage function is to calculate how much better $a_t$ is in $s_t$, Equation ($3$).

$$Q^{\pi}(s_t, a_t) = \sum_{t'=t}^{T} E_{\pi_\theta}[r(s_{t'}, a_{t'})|s_t, a_t] \tag{1}$$

$$V^{\pi}(s_t) = E_{a_t \sim \pi_{\theta(a_t|s_t)}}[Q^{\pi}(s_t, a_t)] \tag{2}$$

$$A^{\pi}(s_t, a_t) = Q^{\pi}(s_t, a_t) - V^{\pi}(s_t) \tag{3}$$

The optimal action $a_t^*$ can be greedily determined by Equation ($4$). If the Q-function $Q^{\pi}(s_t, a_t)$ has a quadratic format, the optimal action can be obtained analytically and easily. Therefore, we design the advantage function in a quadratic format, as in Equation ($5$), which not only incooperate non-linear features but also has a closed-form for the greedy policy as illustrated in Equation ($4$).

$$a_t^* = argmax_a \, Q^{\pi}(s_t, a_t) \tag{4}$$

$$Q^{\pi}(s_t, a_t) = A^{\pi}(s_t, a_t) + V^{\pi}(s_t) \\ = (\mu(s_t) - a_t)^T M(s_t)(\mu(s_t) - a_t) + V(s_t) \tag{5}$$

where the matrix $M(s_t)$ and scalar $V(s_t)$ are outputs from neural networks, and $M(s_t)$ is constrained to be negative-definite in our case for the maximized optimization. The structure of the Q-function is illustrated in Figure 1.

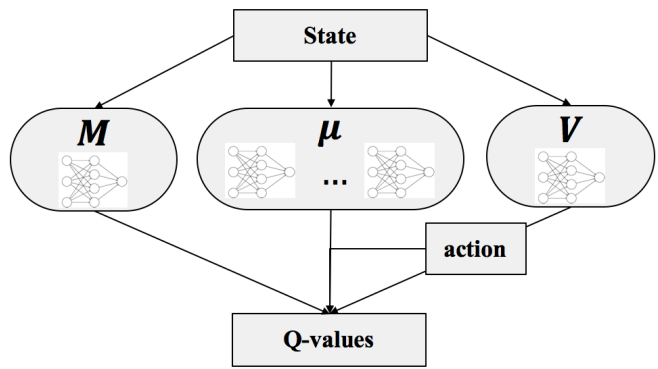

**FIGURE 1 Structure of Q-function approximation.**

From Equation ($5$) it can be observed that $\mu(s)$ plays a critical role in learning the optimal action. If it is designed intuitively as one complicated neural network with multiple layers and thousands even millions of neurons, it may suffer a hard time learning good actions that are meaningful to the driving policy. With this consideration, we use some prior knowledge and design $\mu$ with three neural networks, the outputs from which are considered as pivotal quantities and combined in a defined formula as in Equation (*6*) amd (*7*).

$$a = a_{max} * \tanh\left(\beta_{sen} * a_{tmp}\right) \tag{6}$$

$$a_{tmp} = \frac{\Delta d}{T_{trns}^2} + \Delta v * \frac{\Delta \varphi}{T_{trns}} \tag{7}$$

where $a$ is the lateral action, $a_{max}$ is a learned parameter representing the adaptable maximum acceleration of the driver, $\beta_{sen}$ is a sensitivity factor learned by another neural network, $a_{tmp}$ is a temporary action value obtained from a function of deviation measurements in position, speed, and yaw angle (i.e. $\Delta d$, $\Delta v$, $\Delta \varphi$ respectively), and $T_{trns}$ is a transition time that is the output by the third neural network. Figure 2 depicts the structure of $\mu$.

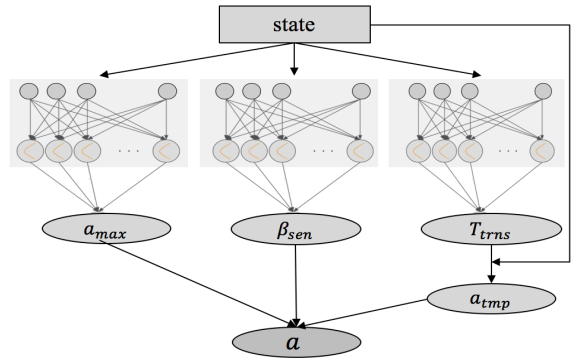

**FIGURE 2 Structure of $\mu$.**

Note that because any smooth Q-function can be Taylor-expanded in this quadratic format near the greedy action, there is not much loss in generality with this assumption if we stay close to the greedy policy in the Q-learning exploration.

*Learning Framework*
There are two parallel loops in the learning procedure, as illustrated in Figure 3. One is a simulation loop where it provides the environment that the vehicle agent interacts with, and the other one is a training loop in which the neural network weights are updated.

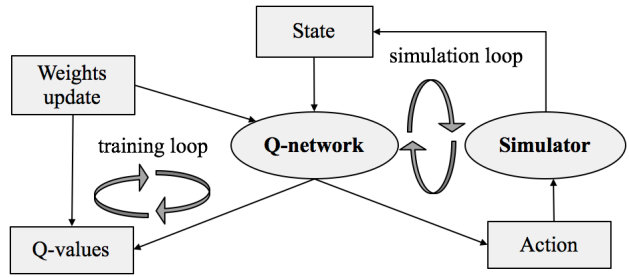

**FIGURE 3 Q-learning structure.**

In the simulation loop, we invoke $\mu(s)$ to obtain the greedy action a for a given state $s$. In order to explore the effect of different actions, the greedy action is perturbed with a Gaussian noise and executed in the simulation as the action taken by the vehicles. After each execution, the new state of the environment $s'$, as well as the reward $r$ under the current state and action, is observed. A tuple of the transition information $(s, a, s', r)$ at each step is recorded in a replay memory $D$.

In the traning loop, samples of tuples are randomly drawn from the replay memory for Q-function weights update. To overcome the inherent instability issues in Q-learning, we use two Q-functions that have the same structure but different sets of parameters, $\theta$ and $\theta^-$, to calculate predicted Q-values ($Q^P$) and target Q-values ($Q^T$) respectively, similar to the experience replay in (*15*). Weights in $\theta$ are updated by gradient descent at every time step, while weights in $\theta^-$ are periodically overwritten with weights in $\theta$. The flow chart in the learning loop is illustrated in Figure 4.

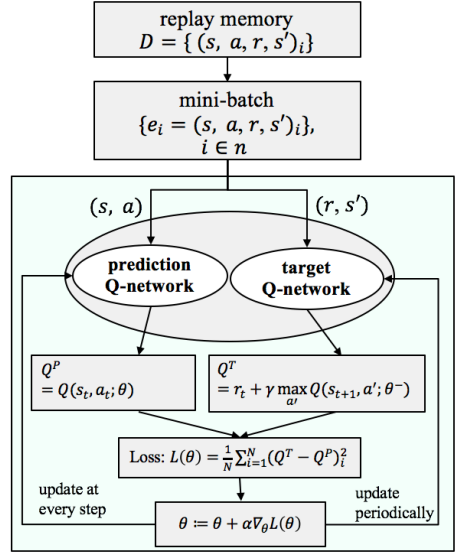

**FIGURE 4 Training procedure with experiment replay.**

Since the training is performed at every time step with a mini-batch randomly drawn from the replay memory $D$, the loss is defined as the errors between predicted Q-values and target Q-values of the mini-batch, as in Equation (8).

$$L(\theta) = \frac{1}{N}\sum_{i=1}^{N}(Q^T - Q^P)_i^2$$
$$= \frac{1}{N}\sum_{i=1}^{N}(r + \gamma max_{a'}Q(s',a',\theta^-) - Q(s,a,\theta))_i^2 \qquad (8)$$

where $i$ is the simulation step, and other variables are as described before.

It is also worth mentioning that the overall learning process is split into two stages, pre-training stage and normal training stage. During the first stage, we only train the neural networks of $M$ and $V$, and keep the initialized parameters of the neural networks in $\mu$ as unchanged. During the second stage, we jointly train all the neural networks and update the parameters simultaneously. This trick helps learn the optimal policy effectively, and leads to better greedy policy.

**Maneuver Control Under Interactive Environment**
With regard to the specific lane-change scenario, the vehicle agent should be able to perform proper longitudinal actions (e.g. car-following behavior) and lateral actions (e.g. shifting to the target lane) while complying with road safety rules and mandatory regulations. These constraints bound the movements in both directions, making the autonomous driving different from problems with freely moving space like drones (*1*) or assistant robots (*16*).

In a lane change maneuver, the longitudinal movement and the lateral movement work collaboratively to complete the task. As car-following controllers have been thoroughly studied for decades and that some off-the-shelf models are ready for use, we resort to a well-developed car-following model, Intelligent Driver Model (IDM) (*17*), with some adaptation to build the longitudinal controller. In contrast, the lateral maneuver can be very complicated in the presence of surrounding vehicles, and hand-crafted formulae are usually far from optimal in controlling the lateral behavior given the uncertainty of other vehicles' future movements. Therefore, we leverage reinforcement learning to learn a lateral maneuver model to integrally cooperate with the longitudinal action and be adaptable to different situations under the dynamic driving environment.

In our work, the outputs from the longitudinal and lateral models are kinematic variables, such as longitudinal acceleration and yaw acceleration. They can be further converted into low-level vehicle dynamics such as throttle, brake, and steering based on the vehicle physical parameters, which is not covered in this paper.

It is worth noting that there can be a hierarchical structure that organizes various functional blocks in an automated driving system. At the higher level, there can be a decision-making module that decides on the strategic actions while at the lower level the control modules generate and execute the commands. In addition, there can also be another functional component, a gap-selection module as used in our study, working in parallel with the two controllers. After the vehicle gets a lane change command from a higher level of the decision-making module, the gap selection module will check the acceptable safety distance between the leader and follower on the target lane based on all the current information (e.g. speed, accelera- tion, position, etc.) of the surrounding vehicles. If the gap is adequate enough to accommodate the speed difference under allowable maximum acceleration/deceleration and to ensure a minimum safety distance under current speed, it is considered as an acceptable gap, and then the lane change controllers will be initiated.

*Longitudinal Maneuver Based on IDM*

Intelligent Driver Model (IDM) is a time-continuous car-following model that describes dynamics of the positions and velocities of single vehicles for the simulation of highway and urban traffic (*17*). Due to space limitation, here we only briefly introduce the modified IDM used in our study.

It is observed that the acceleration phase given by the default IDM in the presence of leading vehicles can be overly conservative resulting in relatively slow speed under moderate traffic conditions. Thus, we make some modifications on the differential equation as in (*9*) for the longitudinal acceleration $a_{lng}$ of the ego vehicle $\alpha$.

$$a_{lng} = \frac{\mathrm{d}v_\alpha}{\mathrm{d}t} = a_m \left(1 - \max\left(\left(\frac{v_\alpha}{v_0}\right)^\delta, \left(\frac{s_0 + v_\alpha T}{s_\alpha} + \frac{v_\alpha \Delta v_\alpha}{2\sqrt{a_m b s_\alpha}}\right)^2\right)\right) \qquad (9)$$

where $\Delta v_\alpha$ is the velocity difference between the ego vehicle $\alpha$ and its preceding vehicle $\alpha - 1$, $v_0$ is the desired velocity of the ego vehicle in free traffic, $s_0$ is the minimum spacing to the leader, $s_\alpha$ is the current spacing, $T$ is the minimum headway to the leader, $a_m$ is the maximum vehicle acceleration, $b$ is the comfortable braking deceleration, and $\delta$ is the exponential parameter. In our study, we set $s_0$ to 5 m by considering the vehicle length, $T$ to 1s, $a_m$ to 2.0 m/s$^2$, $b$ to 1.5 m/s$^2$, and $\delta$ to 4.

Based on the modified model, if the road traffic is sparse and the distance to the leading vehicle $s_\alpha$ is large, the ego vehicle's acceleration is dominated by the free term $a_m(1 - \left(\frac{v_\alpha}{v_0}\right)^\delta)$; in contrast, if the velocity difference is negligible and the distance to the leading vehicle is small, or if the velocity difference is large under high vehicle approaching rates, the ego vehicle $\alpha$ acceleration is approximated by the interaction term $a_m(1 - \left(\frac{s_0 + v_\alpha T}{s_\alpha} + \frac{v_\alpha \Delta v_\alpha}{2\sqrt{a_m b s_\alpha}}\right)^2)$. The modified model helps guarantee reasonable longitudinal behaviors either in free-flow traffic conditions or in limited spacing conditions.

For the scenarios where there are preceding vehicles in both the ego lane and the target lane when the ego vehicle is making a lane change, our IDM car-following model will allow the ego-vehicle to adjust its longitudinal acceleration by balancing between its leaders in both lanes. The smaller value of the two longitudinal accelerations will be used to weaken the potential discontinuity in vehicle acceleration incurred from lane change initiation. Also, the gap selection module will still be working as another safety guard during the whole lane changing process to check whether the gap distance is still acceptable at each time step. If not, the decision-making module will issue a command to alter or abort the maneuver, and the control execution modules will direct the vehicle back to the original lane. In this way, the longitudinal controller takes the surrounding driving environment into account to ensure safety in the longitudinal direction, whereas the lateral controller directs the vehicle to comfortably and efficiently merge into any accepted gap.

*Lateral Maneuver Based on RL*
In this section, we will explain in detail how the reinforcement learning algorithm is applied to the lateral control function.

**Action Space**   To enhance the practicability, we treat the action space as continuous to allow any reasonable real values being taken in the lane change process. Specifically, we define the lateral control action to be the yaw acceleration, $a = \ddot{\theta}$, with the consideration that a finite yaw acceleration ensures the smoothness in steering, where $\theta$ is the yaw angle.

**State Space**   The state information can be in high dimension when images, possibly with a suite of sensor data, are directly taken as input to the reinforcement learning module. Such architectures can be categorized as end-to-end learning models. Alternately, the input state can also be post-processed data in a low dimensional space which only contains the most relevant information to the driving task, e.g. speed, position, acceleration, etc. These data features can be can be extracted by perception modules and/or sensor fusion modules, which are active research topics nowadays in the field of computer vision and perception. Such approaches are generally classified into a modular based learning category.

To explore the learning ability of our designed control module, we choose to turn to the modular based approach. We assume the input data is readily available from embedded vehicle devices, such as GPS, IMU (Initial Measurement Unit), LiDAR, radar, camera, CAN bus, etc., and processed by state-of-the-art data processing modules. The derived state information meets the desired accuracy requirements for the control purpose in our study.

In addition, road geometry also affects the success of a lane change behavior, for example, a curved road segment introduces additional centrifugal force in the lane-changing process. Taking all these into consideration, we define the state space with both vehicle dynamics and road geometry information, as follows.

$$s = (v, a_{lng}, \Delta d_{lat}, \theta, \omega, c) \in S$$

where $v$ is the ego vehicle's velocity, $a_{lng}$ is the longitudinal acceleration, $\Delta d_{lat}$ is the deviation of the lateral position, $\theta$ is the yaw angle, $\omega$ is the yaw rate, and $c$ is the road curvature.

**Immediate Reward Function** The immediate reward is a scalar evaluating the safety, comfortableness and efficiency of the action taken in a given state. In the formulation, we assign negative values to the rewards as as to penalize adverse actions, thus teaching the agent to learn to avoid actions that result in large penalties.

The comfortableness is evaluated by yaw acceleration $a$ and yaw rate $\omega$ as a high yaw acceleration (the absolute value) directly contributes to a large jump in yaw rate that results in significant shifting in the lateral movement. The reward from yaw acceleration is defined as follows.

$$r_{acce} = -w_{acce} \times f_{acce}(a) \tag{10}$$

where $w_{acce}$ is the weight that $r_{acce}$ accounts for in the whole immediate reward $r$, and $f_{acce}$ is a function of $a$ which can be in any format. For our lane change case, we currently define $r_{acce} = -w_{acce}|a|$ because we want to punish high yaw acceleration.

The reward from yaw rate is given below.

$$r_{rate} = -w_{rate} \times f_{rate}(\omega) \tag{11}$$

where $w_{rate}$ is the weight of $r_{rate}$ in $r$, and $f_{rate}$ is a function for evaluating $\omega$. We use $r_{rate} = -w_{rate} \times |\omega|$ in our study to punish high yaw rate.

Safety and efficiency are evaluated by a variable denoted as lateral position deviation, $\Delta d_{lat}$. It is considered that the larger the deviation is the longer the time the vehicle consumes to finish the lane change. Additionally, a large deviation also increases the risks of accidents in driving. Furthermore, safety is also considered in longitudinal control by the IDM and the gap selection module. The reward, $r_{dev}$, from lateral deviation is calculated as follows.

$$r_{dev} = -w_{dev} \times f_{dev}(\Delta d_{lat}) \tag{12}$$

where $w_{dev}$ is the weight of $r_{dev}$ in $r$, and $f_{dev}$ is a function of $\Delta d_{lat}$. In our study, we design $f_{dev}$ as the proportion of the current lateral deviation $\Delta d_{lat}$ over the average lateral deviation $d_{avg}$ during a lane change which we take as half of a lane width. Therefore, $r_{dev} = -w_{dev}(\Delta d_{lat}/d_{avg})$.

The weighting parameters, $w_{acce}$, $w_{rate}$, and $w_{dev}$, are hyperparameters and manually tuned through multiple training episodes. They are set to 2.0, 0.5, 0.05 in the lane change case for the best performance.

The immediate reward $r$ in a single step is the summation of the three parts. In the evaluation of the RL algorithm, the total reward $R$ is a primary evaluation indicator which is an accumulated value from immediate rewards $r$ over the lane changing process. Equally, $R$ can also be viewed as a composition of three sub-total rewards, donated as $R_{aace}$, $R_{rate}$, and $R_{dev}$ , and expressed as in Equation (*13*).

$$R = \sum_{k=1}^{N}(r_{acce} + r_{rate} + r_{dev})_k = \sum_{i=1}^{N}(r_{acce})_i + \sum_{i=1}^{N}(r_{rate})_k + \sum_{i=1}^{N}(r_{dev})_k \tag{13}$$

where $k$ is the time step in a lane-change process.

## SIMULATION AND RESULT

We test our proposed algorithms through a customized simulation platform where a learning agent is able to, on one hand, interact with the driving environment and improve it driving behavior by trials and errors, and on the other hand the surrounding vehicles can also dynamically respond to the ego vehicle's real-time action, which is currently not provided by most simulation software.

**Simulation Settings**

The simulation scenario is a highway segment with three lanes on each direction. All lanes have the same lane width of 3.75m, and the overall length of the testing track is 1000 m. The simulated traffic can be customized to generate diverse driving conditions. For example, the initial speed, the departure time, and the speed limit of each individual vehicle can all be set to random values as long as they are within reasonable ranges. Randomly selected vehicles in the middle lane can get lane change commands (change left or change right) after travelling for about 150 meters. Vehicles on the other lanes stay on its own lane and can surpass or yield to the intended lane-changing vehicles. All vehicles can perform practical car-following behaviors with our adapted IDM. Additionally, aggressive drivers can be simulated with high acceleration profile and close car-following distance, and conversely for defensive drivers.

In particular, we set the departure time interval to a range of 5s-10s, the individual initial speed to a range of 30 km/h-50 km/h, and the achievable speed limits of individual vehicles to a range of 80 km/h-120 km/h, to generate diverse driving conditions. An illustrative scene of the simulation scenario is shown in Figure 5 where the red vehicle is the lane-changing ego vehicle, and the yellow vehicles are the surrounding vehicles.

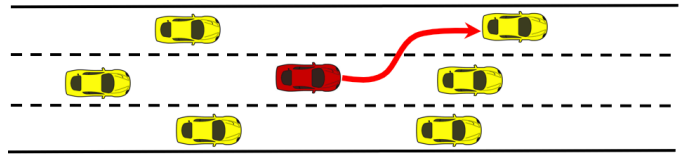

**FIGURE 5 Illustration of simulated lane change scenario.**

**Training Results**

In our training, the simulation time interval $dt$ is set to 0.1s, the learning rate is set to 0.0005, the discount factor $\gamma$ is 0.95, the batch size is 64, the Q-network update rate is every 1000 steps. The pre-train step is set to 200,000, during which one third, around 6500 vehicles, on the middle lane performed lane change maneuvers. Multiple training steps are used, as listed in [40,000, 80,000, 120,000, …, 400,000], to save intermediate models for driving performane comparison in the validation phase.

In the training, we recorded the mini-batch loss as defined in Equation (*8*) at every 20 steps, as well as the total rewards as defined in Equation (*13*) of each lane-changing vehicle whenever it finished a lane change maneuver. Figure 6(a) shows the loss curve, and Figure 6(b) shows the evolutionary graphs of the total reward $R$ gathered from lane-change vehicles.

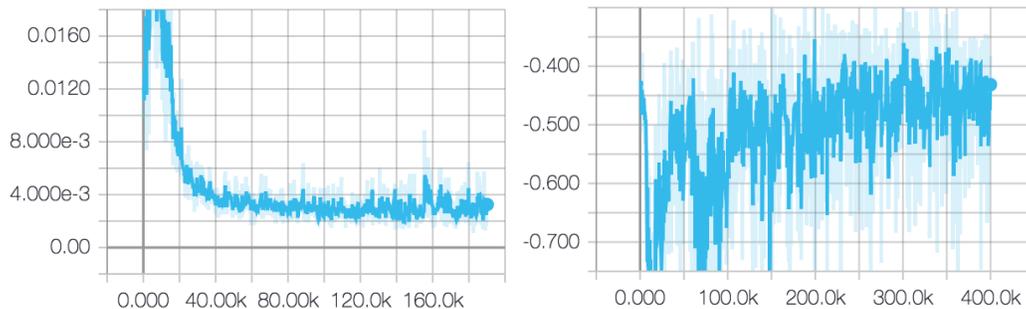

**FIGURE 6  (a) Loss accumulated in training.    (b) Total rewards of individual vehicles in training.**

From Figure 6(a) we can observe that the training loss curve shows an obvious convergence along with training steps, which means the predicted Q-values are quite close to the target Q-values. The total reward graph in Figure 6(b), which is the primary indicator of the RL performance, further proves the convergence. It shows that the negative values of the total rewards go up and reach to a roughly steady level over the training, indicating that vehicles have learned to change lane with smaller penalties (small absolute values).

Since each point in the reward curve represents only one random individual vehicle's performance under the corresponding training parameters, it might not be confidently enough to say the vehicle has learned a good driving policy. Therefore, we conducted additional tests on the vehicle driving performance. For the 10 saved models, we freeze their parameters and run around 100 lane-changing vehicles on each saved model, and then calculated their averaged total reward as the evaluator for each model. The averaged total reward curve is shown in Figure 7. It is obvious that as the training goes on, the averaged total rewards gained during the lane changing processes increase steadily.

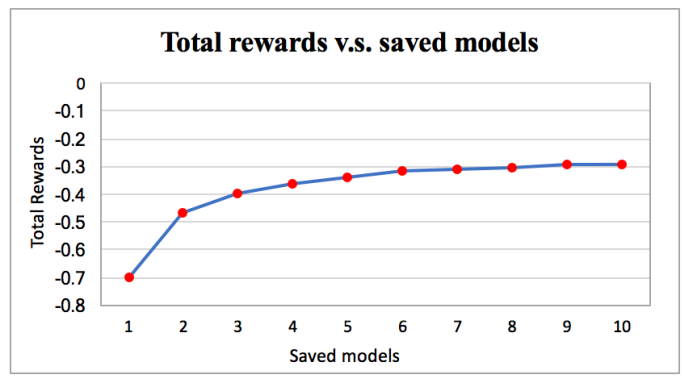

**FIGURE 7 Averaged total reward on saved models during training.**

To further verify the lane-changing behavior did get improved by our proposed RL algorithm, we also compared the driving performance of some randomly selected lane-change vehicles under two saved models, a model with 40,000 training steps and a model with 400,000 training steps. Figure 8 demonstrates the curves of yaw acceleration and yaw rate.

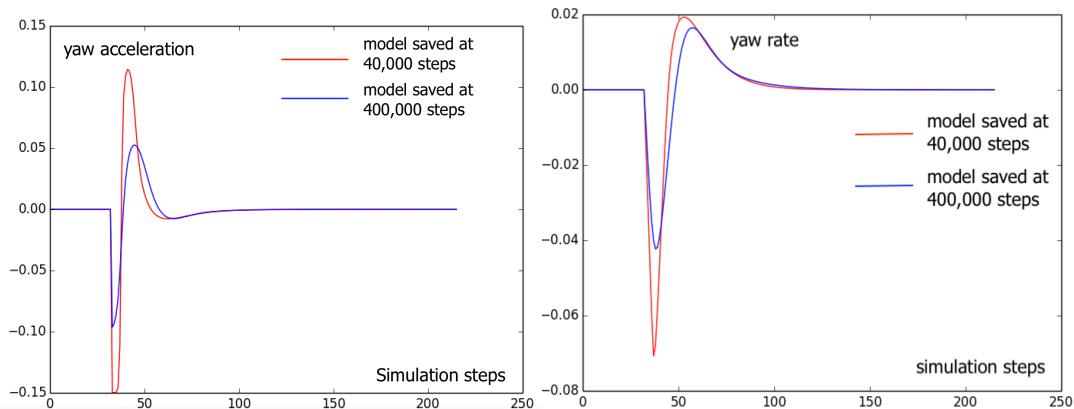

**FIGURE 8 (a) Yaw acceleration under two models.      (b) Yaw rate under two models.**

The blue curves in Figure 8 (a) and (b), representing the performance of the model saved at the end of the training (i.e. 400,000 steps), change moderately compared with the red curves generated by the naïve model at roughly the beginning of the training (i.e. 40,000 steps). It proves that with the training by RL, the vehicle agent learns to change lane in a desirable way.

**CONCLUSION AND DISCUSSION**

In this work, we designed a quadratic format of Q-function approximator based on normalized advantage function and applied it to learn the lane-change behavior under interactive driving environment. The state space and action space are both treated as continuous to learn practical driving maneuvers. The closed form of the quadratic function saves computation costs as well as the efforts of designing complicated policy network.

Covergence in training loss and total rewards, as well as improved driving performance indicat the capability of the vehicle in learning the defined policy, i.e. a safe, comfortable, and efficient lane change policy. This also demonstrates the promising aspects of applying reinforcement learning in solving other related autonomous driving problems.

To further extend our research, we will design to learn the longitudinal action and lateral action together in a two-dimensional action space and use sensory data of different modality as input in a high dimensional state space. In addition, concatenating state information from successive steps or resorting to LSTM (Long Short-Term Memory) algorithms to learn the state representation is worth trying. Also, formulating different types of the reward function or directly learning it from driving trajectory demonstrations is promising to further improve the driving performance. Furthermore, another meaningful attempt is to combine the reinforcement learning and traditional control methods (e.g. MPC) to make the best of both approaches where the reinforcement learning algorithm is used to learn a desirable trajectory reference, and the traditional optimization-based controller is used to quickly calculate a reliable control command for the vehicle to follow. These are all promising aspects worth pursuing.

**AUTHOR CONTRIBUTION STATEMENT**

The authors confirm contribution to the paper as follows: study conception and design: Pin Wang and Ching-Yao Chan; data collection: Pin Wang; analysis and interpretation of results: Pin Wang and Ching-Yao Chan; draft manuscript preparation: Pin Wang and Ching-Yao Chan. All authors reviewed the results and approved the final version of the manuscript.